\documentclass[lettersize,journal]{IEEEtran}
\usepackage{amsmath,amsfonts}
\usepackage{algorithmic}
\usepackage{array}
\usepackage[caption=false,font=normalsize,labelfont=sf,textfont=sf]{subfig}
\usepackage{textcomp}
\usepackage{url}
\usepackage{verbatim}
\usepackage{graphicx}
\usepackage{bbding} 
\usepackage{booktabs}
\usepackage{cite}
\usepackage{multirow}
\usepackage{color}
\usepackage{bm}
\usepackage{makecell}
\usepackage{colortbl}
\usepackage{url}
\definecolor{lightgray}{rgb}{0.7,0.7,0.7}

\begin{document}
\title{Noise-Informed Diffusion-Generated Image Detection with Anomaly Attention}
\author{Weinan Guan, Wei Wang, Bo Peng, Ziwen He, Jing Dong, Haonan Cheng
        % <-this % stops a space
\thanks{
Weinan Guan is with the School of Artificial Intelligence, University of Chinese Academy of Sciences, Beijing 100049, China, and also with New Laboratory of Pattern Recognition (NLPR), Institute of Automation, Chinese Academy of Sciences (CASIA), Beijing 100190, China (E-mail: weinan.guan@cripac.ia.ac.cn).

Wei Wang (corresponding author), Bo Peng and Jing Dong are with New Laboratory of Pattern Recognition (NLPR), Institute of Automation, Chinese Academy of Sciences (CASIA), Beijing 100190, China (E-mail: wwang@nlpr.ia.ac.cn; bo.peng@nlpr.ia.ac.cn; jdong@nlpr.ia.ac.cn).

Ziwen He is with Nanjing University of Information Science and Technology (E-mail: ziwen.he@nuist.edu.cn).

Haonan Cheng is with State Key Laboratory of Media Convergence and Communication, Communication University of China, Beijing 100024, China (E-mail: haonancheng@cuc.edu.cn).
}}

\markboth{Journal of \LaTeX\ Class Files,~Vol.~18, No.~9, September~2020}%
{How to Use the IEEEtran \LaTeX \ Templates}

\maketitle

\begin{abstract}
With the rapid development of image generation technologies, especially the advancement of 
Diffusion Models, 
the quality of synthesized images has significantly improved, raising concerns among researchers about information security. To mitigate the malicious abuse of diffusion models, diffusion-generated image detection has proven to be an effective countermeasure.
However, a key challenge for forgery detection is generalising to diffusion models not seen during training. In this paper, we address this problem by focusing on image noise. We observe that images from different diffusion models share similar noise patterns, distinct from genuine images. Building upon this insight, we introduce a novel Noise-Aware Self-Attention (NASA) module that focuses on noise regions to capture anomalous patterns. 
To implement a SOTA detection model, we 
incorporate NASA into Swin Transformer, forming an novel detection architecture NASA-Swin. Additionally, we employ a cross-modality fusion embedding to combine RGB and noise images, along with a channel mask strategy to enhance feature learning from both modalities. Extensive experiments demonstrate the effectiveness of our approach in enhancing detection capabilities for diffusion-generated images. When encountering unseen generation methods, our approach achieves the state-of-the-art performance.
Our code is available at \textcolor{magenta}{\textit{\url{https://github.com/WeinanGuan/NASA-Swin}}}.
\end{abstract}


\section{Introduction}
\IEEEPARstart{R}{ecently}, the computer vision community has witnessed remarkable progress in image synthesizing methods \cite{rombach2022high, dhariwal2021diffusion, liu2022pseudo, nichol2021improved, song2020denoising, nichol2021glide, gu2022vector} due to the strong image generation capability of 
Diffusion Models, such as
Denoising Diffusion Probabilistic Models (DDPMs) \cite{ho2020denoising} 
and Denoising Diffusion Implicit Models (DDIMs) \cite{song2020denoising}. 
The emergence of numerous diffusion models provides a range of alternative image generation methods, enabling the production of photorealistic images. 
Particularly, the text-to-image diffusion models, like Stable Diffusion \cite{rombach2022high}, DALL$\cdot$E2 \cite{ramesh2022hierarchical}, Midjourney \cite{Midjourney} and Imagen \cite{saharia2022photorealistic}, only require a phase or sentence to generate the desired images. Compared to the previous GAN-based models, the quality of generated images and convenience of generation procedures have been further improved.
However, this also raises potential security concerns. For instance, attackers can easily exploit these approaches for malicious purposes, such as spreading fake news, developing new deepfake techniques to defame celebrities and falsifying evidence. Therefore, it is imperative to develop forgery detectors for images generated by diffusion models.

In this study, our primary focus is on developing an effective detector for diffusion-generated images. Previous forgery detection methods are primarily designed to detect GAN-generated (Generative Adversarial Networks \cite{goodfellow2014generative}) images \cite{zhang2019detecting, wang2020cnn, yu2019attributing}. However, diffusion models employ a distinct generative process compared to previous generators, leading to the ineffectiveness of traditional detection methods  \cite{wang2023dire}. While forgery detection may appear as a straightforward image classification task, simple binary classifiers for distinguishing diffusion-generated from genuine images are often proved ineffective when dealing with images generated by previously unseen diffusion models \cite{wang2023dire, zhu2023genimage}.

\begin{figure}[t]
  \centering
  \includegraphics[width=\linewidth]{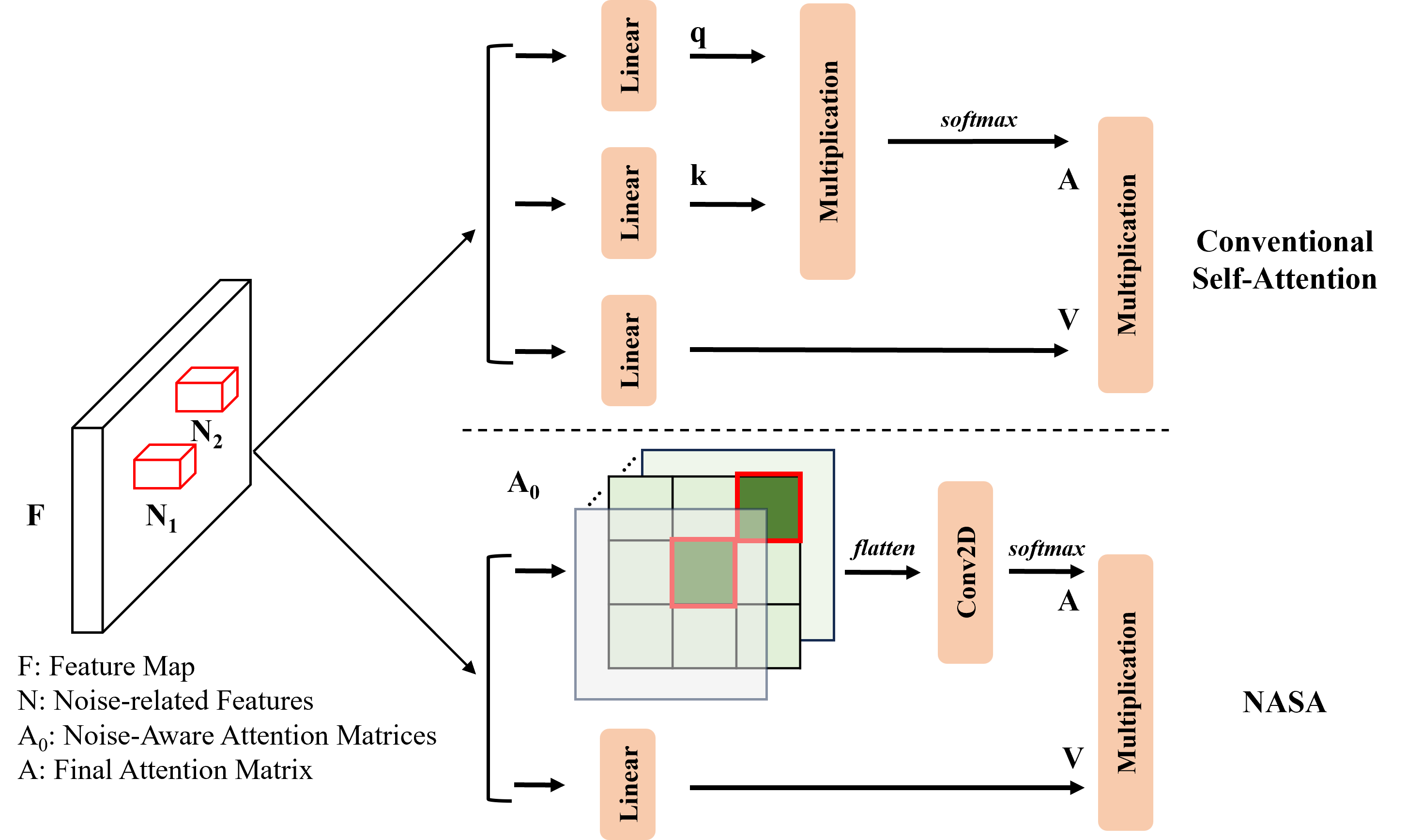}

   \caption{The comparison between NASA and the conventional self-attention mechanism. NASA allocates more attention to the noise-related features.}
   \label{fig:NASA-Tradition-Compare}
\end{figure}

\begin{figure}[t]
  \centering
  \includegraphics[width=\linewidth]{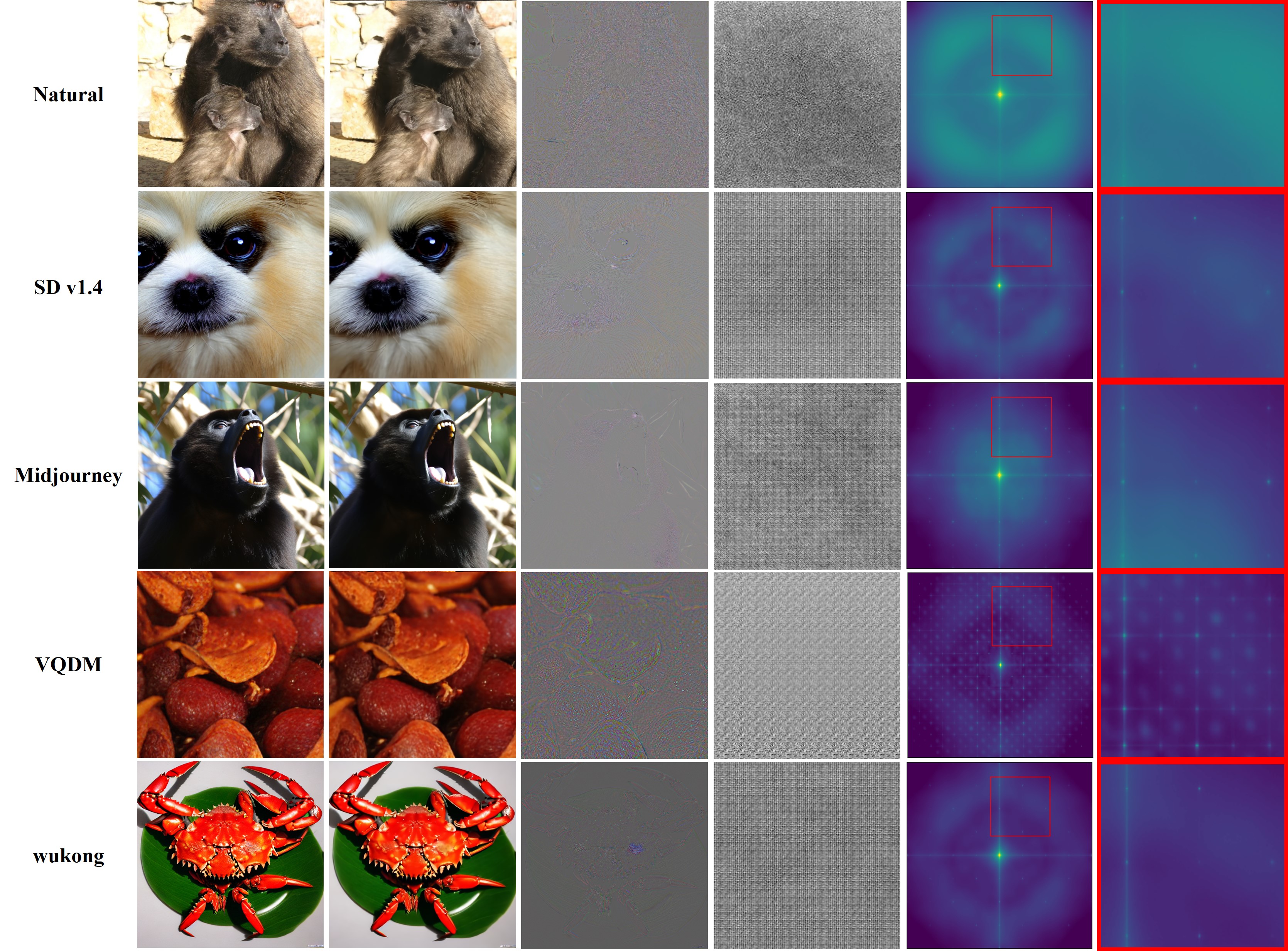}

   \caption{{The denoised images and their respective noise residuals are computed using RIDNet \cite{anwar2019real}. The first column consists of the genuine and generated images sampled from GenImage \cite{zhu2023genimage}, the second column displays their denoised versions, and the third column exhibits their corresponding noise residuals. The fourth column provides visualizations of the average image noise residuals of $20,000$ images from the same data source, indicating the differences between generated and natural images in the spatial domain (i.e. the structured grid textures in generated image noise). The last two columns respectively display the average frequency spectra of $20,000$ image noise residuals from the same data source and the enlarged high-frequency regions of each spectrum, highlighting the distinct frequency responses between generated and natural images (the high-frequency spots in generated image spectra).}}
   \label{fig:RID-Residual-FFT}
\end{figure}

In this paper, we explore the detection clues within diffusion-generated images by closely investigating the image noise. 
We observe that genuine and generated image noise exhibits distinct {spatial patterns and high-frequency responses. As shown in Fig~\ref{fig:RID-Residual-FFT}, the image noise from various diffusion models displays similar spatial and frequency patterns, such as structured grid textures in spatial domain and prominent highlights in high-frequency regions. In contrast, such structured grid texture patterns in the spatial domain and the pronounced high-frequency features are notably absent in nature image noise.}
This phenomenon may be attributed to anomalous noise patterns arising from the denoising process across all diffusion models.
This inspires us to utilize the residual noise to enhance the generalization ability of forgery detection.

To harness the potential of image noise 
in forgery detection, we introduce a novel detection architecture based on an advanced Transformer model (i.e. Swin Transformer \cite{liu2021swin}). To enhance the focus on noise-related information, we design a noise-aware self-attention (NASA) module for attention calculation in the self-attention mechanism. The image noise may lead to anomalous activation patterns (i.e. anomalous feature values) in feature maps. Unlike the conventional self-attention module, NASA allocates more attention to the noise-related features. The comparison is depicted in Fig~\ref{fig:NASA-Tradition-Compare}. This NASA-based Transformer block is employed to construct an additional branch in the intermediate stages of our model. Besides, we employ a cross-channel data fusion strategy and develop a hierarchical embedding method to integrate RGB and noise residual images as the input. A channel mask strategy is utilized to encourage the model to learn complementary features from both modalities.

We conduct extensive experiments to demonstrate that the inclusion of noise residuals significantly improves the model's generalization ability. Our results indicate that our method achieves superior detection performance on generated images from previously unseen diffusion models. When compared to previous detectors, our framework notably surpasses the current state-of-the-art methods. Our main contributions are three-fold as follows.

\begin{itemize}
    \item We provide a novel perspective on detecting diffusion-generated images by utilizing the anomalous noise patterns and propose a transformer based model with a carefully designed Noise-Aware Self-Attention module.

    \item We propose a cross-modality fusion embedding module to integrate RGB and noise residual images. And a channel mask strategy is employed to encourage the model to learn complementary features from both inputs.

    \item Extensive experiments demonstrate that the proposed method achieves state-of-the-art performance in diffusion-generated image detection.
\end{itemize}

\section{Related Work}
\textbf{Diffusion Models.} Inspired by \cite{sohl2015deep}, Denoising Diffusion Probabilistic Model (DDPM) \cite{ho2020denoising} provides a novel generative paradigm, and achieves a competitive performance compared to PGGAN \cite{karras2017progressive}. DDPM gradually transforms noise into a coherent image using a series of learnable reverse diffusion steps. It addresses the challenge of generating high-quality, realistic images from noise, using a Markov chain to progressively denoise the image. Compared to DDPM, DDIM \cite{song2020denoising} enhances sampling efficiency by modifying the sampling process to be non-Markovian, allowing each step to reference earlier states directly. DDPM and DDIM serve as foundational models for diffusion-based generation, forming the basis for numerous related works in the field.

Recent diffusion models can be broadly categorized into two groups: unconditional generation models and conditional generation models. The former primarily emphasize image quality \cite{hong2023improving, tanveer2023diffusion, dhariwal2021diffusion, rombach2022high} and inference efficiency \cite{tang2023deediff, song2020score, karras2022elucidating}. In contrast, the latter leverage text to guide the image generation process \cite{nichol2021glide, kim2022diffusionclip, zhang2023sine, li2023guiding}. 
In practice, these text-to-image diffusion models are extremely popular among people due to their convenience of generation procedures and exceptional image generation performances, such as Stable Diffusion \cite{rombach2022high}, Midjourney \cite{Midjourney}, VQDM \cite{gu2022vector} and Wukong \cite{Wukong}.
{The success of diffusion models has recently expanded to video generation, with several advanced text-to-video approaches emerging. Sora \cite{Sora}, a text-to-video generative AI model, can produce up to one-minute videos with strong temporal consistency while maintaining adherence to text instructions \cite{liu2024sorareviewbackgroundtechnology}. With advanced understanding of natural language and visual semantics, Veo \cite{Veo} can generate a video that closely follows the prompt. It accurately captures the nuance and tone in a phrase, rendering intricate details within complex scenes. Additionally, Gen-3 Alpha \cite{Gen-3} excels at generating expressive human characters with a wide range of actions, gestures, and emotions, and CogVideoX \cite{yang2024cogvideox} can generate 10-second continuous videos with a frame rate of $16$ fps and resolution of $768\times1360$ pixels.}
Although these models provide novel visual experiences for users, they raise several information security issues. Thus, it is crucial to develop forgery detection methods to defend malicious attacks.

\begin{figure*}[t]
  \centering
   \includegraphics[width=\linewidth]{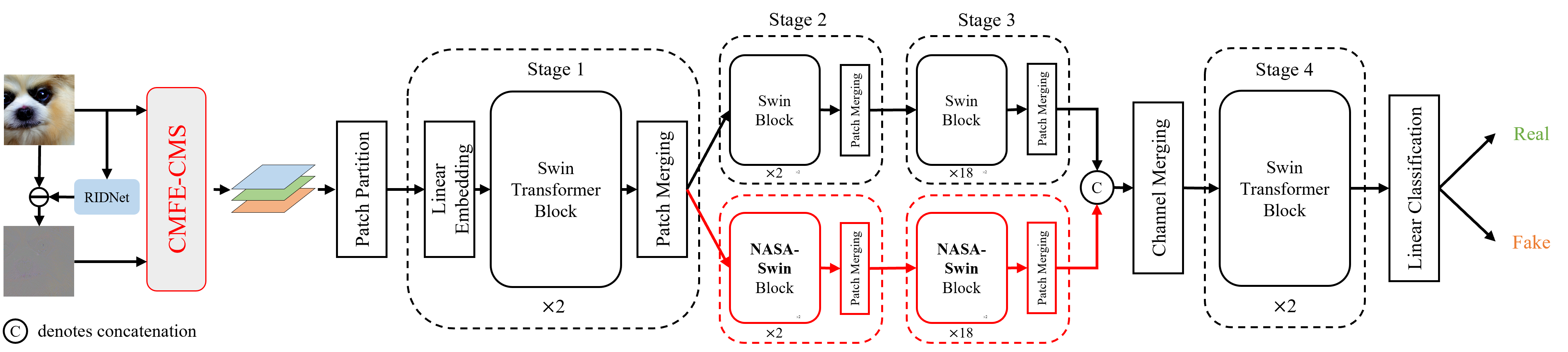}

   \caption{Overview of our framework. {CMFE-CMS denotes the proposed Cross-Modality Fusion Embedding (Section~\ref{sec:CMFE}) and Channel Mask Strategy (Section~\ref{sec:CMS}), which are utilized to process the input pairs of RGB and noise images. In contrast to the standard Swin Transformer, our method incorporates an additional noise-aware branch (highlighted in red) comprising NASA-Swin Blocks to capture noise-related information in the intermediate stages.}}
   \label{fig:architecture}
\end{figure*}

\textbf{Generated Image Detectors.} Over the past few years, detecting generated images has attracted significant attention. As a binary classification problem, several image classifiers, including ResNet \cite{he2016deep}, Xception \cite{chollet2017xception}, EfficientNet \cite{tan2019efficientnet}, and Swin Transformer \cite{liu2021swin}, have served as baseline models \cite{rossler2019faceforensics++, dolhansky2020deepfake, 9484407, cao2022end, dong2023implicit}. Additionally, some approaches have focused on hand-crafted features, such as saturation cues \cite{mccloskey2019detecting}, co-occurrence matrices \cite{nataraj2019detecting}, and physiological signals \cite{matern2019exploiting, yang2019exposing, qi2020deeprhythm}. However, these methods often struggle to generalize effectively due to the continuous innovation in generative techniques. To address this, several well-designed methods have been proposed to develop new detection cues that enhance generalization. These cues include spatial artifacts \cite{li2020face, zhao2021learning, shiohara2022detecting, fei2022learning, 10054130, 9505637}, frequency artifacts \cite{qian2020thinking, liu2021spatial, dong2022think, chen2021local, 10004978, 9854878}, and semantic clues \cite{mittal2020emotions, haliassos2021lips, dong2022protecting, haliassos2022leveraging, 9694644}. Another approach \cite{wang2020cnn} has achieved a promising performance through carefully designed data and data augmentation strategies.

With the emergence of diffusion models, efforts have gradually shifted toward detecting diffusion-generated images. DIRE \cite{wang2023dire} represents an early exploration of this task, focusing on the distinctions in image content reconstruction between genuine and generated images. The authors leverage the reconstruction error of images for detection{, discovering that diffusion-generated images can be more accurately reconstructed by the same diffusion model compared to real images}. 
Following this, SeDID \cite{ma2023exposing} explores the intermediate steps in the diffusion and reverse diffusion processes, {combining both statistical and neural network-based methods to address the diffusion-generated image detection task. By utilizing the error between the reverse sample and the denoise sample at a specific timestep, SeDID exhibits superior performance in detecting diffusion-generated images.} 
{However, both DIRE and SeDID require multi-step DDIM sampling processes, resulting in low efficiency for real-world applications \cite{luo2024lare}. To tackle this problem, LaRE$^2$ \cite{luo2024lare} proposes to reconstruct feature error in the latent space for generated image detection.}
{FIRE \cite{chu2024fire} investigates the influence of frequency decomposition on reconstruction error, observing that diffusion models struggle to accurately reconstruct mid-band frequency information in real images. Based on this limitations, FIRE provides a robust method for detecting diffusion-generated images.}
{Besides the reconstruction-based methods}, MultiLID \cite{lorenz2023detecting}, a variant of LID \cite{ma2018characterizing} operating in lower-dimensional spaces, {analyzes the differences of various local features to capture forgery patterns in generated images, demonstrating} promising detection performance.
{UniFD \cite{ojha2023towards} explores the benefits of pre-trained models in the synthetic image detection task, resorting to their prior knowledge to improve the detection performance. It combines frozen CLIP-ViT with a learnable linear layer for identifying the genuine and generated samples, achieving surprisingly good generalization ability.}
{Furthermore, Fatformer \cite{liu2024forgery} develops a forgery-aware adapter to adapt image features to discern and integrate local forgery traces within image and frequency domains, obtaining well-generalized forgery representations.}

In our method, we design a novel self-attention mechanism to leverage the image noise for detection.
Especially, it may also capture the artifacts in GAN-generated images, which often exhibit anomaly patterns in high frequency \cite{wang2020cnn}.
We further develop a novel detection architecture for generated image detection, resulting in the state-of-the-art performance.

\section{Method}
\subsection{Motivation}
\textbf{Generation pipeline.} To explore the common forgery artifacts among the images created by different diffusion models, we first review the typical generation pipeline. 
The Diffusion model training procedure can be roughly divided into two stages, i.e., adding noise and denoising. First, Gaussian noise is gradually added to the real images. And then, a U-Net \cite{ronneberger2015u, ho2020denoising} is trained to gradually recover the original images by predicting the added noise in every step. In the generation procedure, a realistic image can be generated from a given noise image by the denoising process.

\noindent

\textbf{Artifacts brought by denosing processes.} Though diffusion models can generate visually clear and realistic images, the denoising procedure does not explicitly ensure that the generated images share consistent image noise patterns with the genuine images. Assuming that the generated images have anomalous noise patterns, we can utilize these discrepancies to assist in forgery detection.

To substantiate this assertion, we conduct experiments designed to uncover forgery artifacts within image noise residuals. RIDNet \cite{anwar2019real} has been empirically demonstrated to be an exceptionally efficient and flexible denoising model, adapted for handling both spatially variant and invariant noise, regardless of the noise standard deviation. We employ RIDNet to generate a noise-reduced version of the image. The image's noise residuals are then derived through the difference between the original image and the denoised version, as shown in Fig~\ref{fig:RID-Residual-FFT}. 
Nevertheless, it is a challenging endeavor to identify concealed forgery artifacts in the image noise residuals 
{by directly observing a single noise image. Inspired by }
the previous works \cite{zhu2023genimage, wang2020cnn}, we separately sample $20,000$ images for each image source (different generators or natural images).  
{Subsequently, we respectively visualize the average noise patterns and the average frequency spectra of their image noise residuals in order to analyze the artifacts generated by diffusion models. We would like to investigate whether the generated images possess similar average noise patterns and average frequency spectra to those of natural images. As shown in Fig~\ref{fig:RID-Residual-FFT}, it is evident that the average noise pattern and the average spectral profile of the noise residuals in generated images display noticeable distinction from those of natural images (e.g. SD v1.4 exhibits structured grid textures in the spatial domain and prominent high-frequency highlights in the frequency domain).}
This observation reveals that the noise residuals in images hold crucial information that can be exploited to distinguish between genuine and generated images. This information can be attributed to the anomalous image noise patterns due to the image generation process.

\begin{figure}[t]
  \centering
   \includegraphics[width=\linewidth]{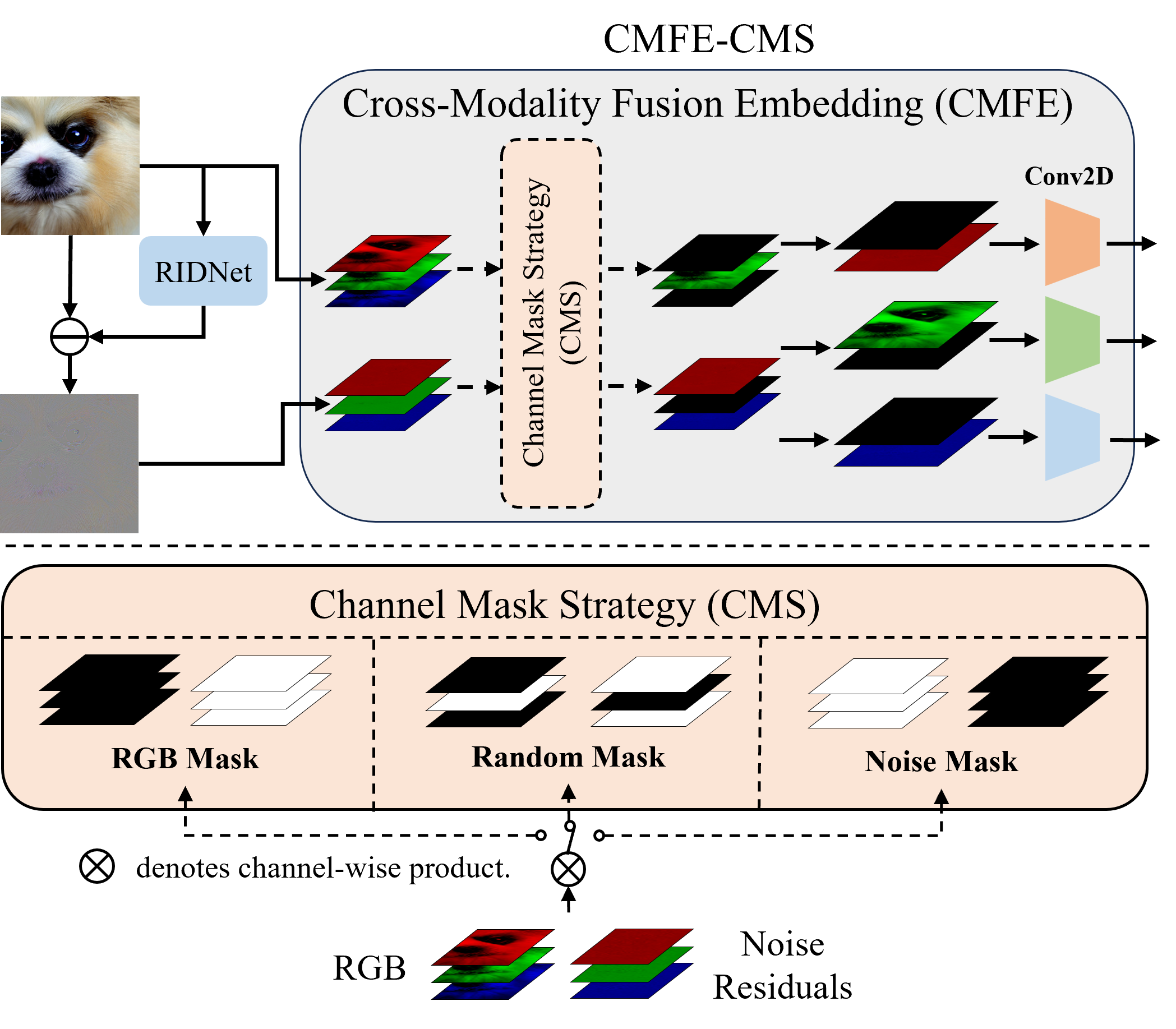}

   \caption{The illustration of Cross-Modality Fusion Embedding (CMFE, top) module and Channel Mask Strategy (CMS, bottom). The dotted lines and boxes represent optional operations, providing choices for selection.}
   \label{fig:CMFE-CMS}
\end{figure}

\subsection{Overall Architecture}
The analysis presented above motivates us to delve deeper into the noise residuals to improve forgery detection performance. To this end, we focus on integrating information related to noise residuals into the detection architecture. 
We argue that the noise residuals present in the generated images exhibit inconsistent noise patterns with genuine images due to the denoising procedures, which serves as crucial detection clues.

An overview of our framework is presented in Fig~\ref{fig:architecture}. We begin the process by employing the proposed \textbf{Cross-Modality Fusion Embedding (CMFE)} module and \textbf{Channel Mask Strategy (CMS)} to fuse data from both modalities (more details in Sec~\ref{sec:CMFE} and Sec~\ref{sec:CMS}), thereby extracting unified and informative representations in the initial stage. Subsequently, we introduce an additional noise-aware branch in the middle stages, comprising several Transformer blocks with modified self-attention computation (\textbf{Noise-Aware Self-Attention} module, \textbf{NASA}) (more details in Sec~\ref{sec:NASA}). The NASA-based Transformer blocks are leveraged to independently capture noise-related information. Following this, we employ a Channel Merging Module composed of linear layers to fuse the features from both branches. The final stage, consisting of standard Transformer blocks, incorporates a linear classification layer to yield detection results.

\subsection{Cross-Modality Fusion Embedding} \label{sec:CMFE}

To harness the full potential of image noise residuals, we first devise an effective fusion strategy for integrating pairs of RGB and residual noise images. Taking inspiration from Depthwise Separable Convolutions in Xception \cite{chollet2017xception}, we propose a Cross-Modality Fusion Embedding \textbf{(CMFE)} strategy, as illustrated in Fig~\ref{fig:CMFE-CMS}. In this method, images from both modalities are interleaved by channels, rather than being directly concatenated. The Channel Mask Strategy shown in Fig~\ref{fig:CMFE-CMS}, is a unique data augmentation strategy designed to enhance the performance of our model, as explained in Sec~\ref{sec:CMS}. Following this, we partition the fused data into three groups based on their {color channels} and subsequently apply separate convolution operations within each group. Similar to Depthwise Convolution, our design enhances the information within each color channel with image noise residuals while preserving the relative independence of these channels. This ensures efficient and organized information exchange across various modalities and channels.

\begin{figure}
    \centering
    \subfloat[\footnotesize A normal image matrix.]{\includegraphics[width=.45\linewidth]{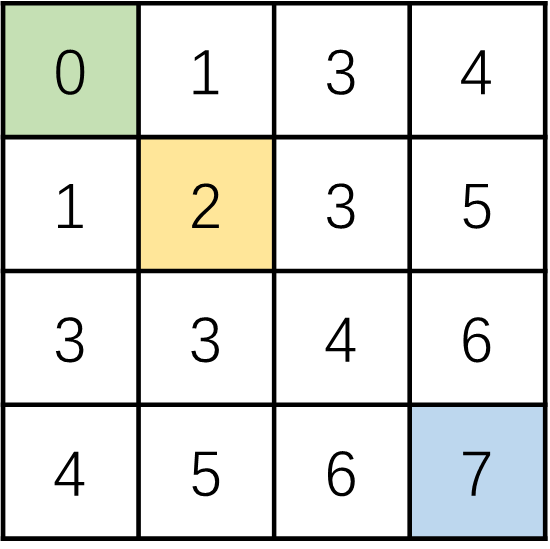}
    \label{fig:NASA-normal-matrix}}
    \hfill
    \subfloat[\footnotesize An image matrix with noise.]{\includegraphics[width=.45\linewidth]{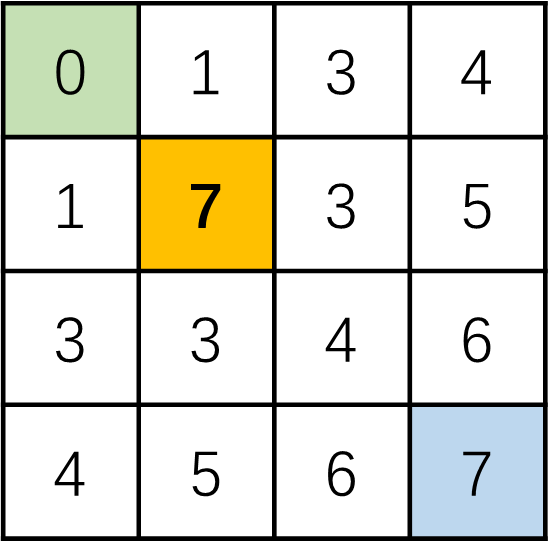}
    \label{fig:NASA-noise-matrix}}
    \hfill
    \caption{A simple example for comparing an feature matrix with and without noise. The green elements represent reference points, while the orange elements in subfigure (a) and (b) correspond to normal and noise-related features, respectively. The blue elements show large differences in feature values relative to the reference points, due to the image content variation.}
    \label{fig:NASA-Matrix-Interpre}
\end{figure}

\subsection{Channel Mask Strategy} \label{sec:CMS}
To maximize the effectiveness of our method, we introduce a Channel Mask Strategy \textbf{(CMS)}. Specifically, we provide {four} unique mask options {(i.e. image vs. noise masks, random channel mask, and no mask)}, randomly selected during the training phase. The details of our mask strategy is demonstrated in Fig~\ref{fig:CMFE-CMS}. It is based on two core designs: 1) The modality (RGB or Noise) mask, which randomly conceals one modality image, serves the purpose of preventing the model from solely learning forgery artifacts from a single modality. 2) The random channel mask, resembling the concept of Cutout \cite{devries2017improved}, operates on channels, encouraging the network to prioritize complementary and less prominent features from all channels.

\begin{figure}[t]
  \centering
   \includegraphics[width=\linewidth]{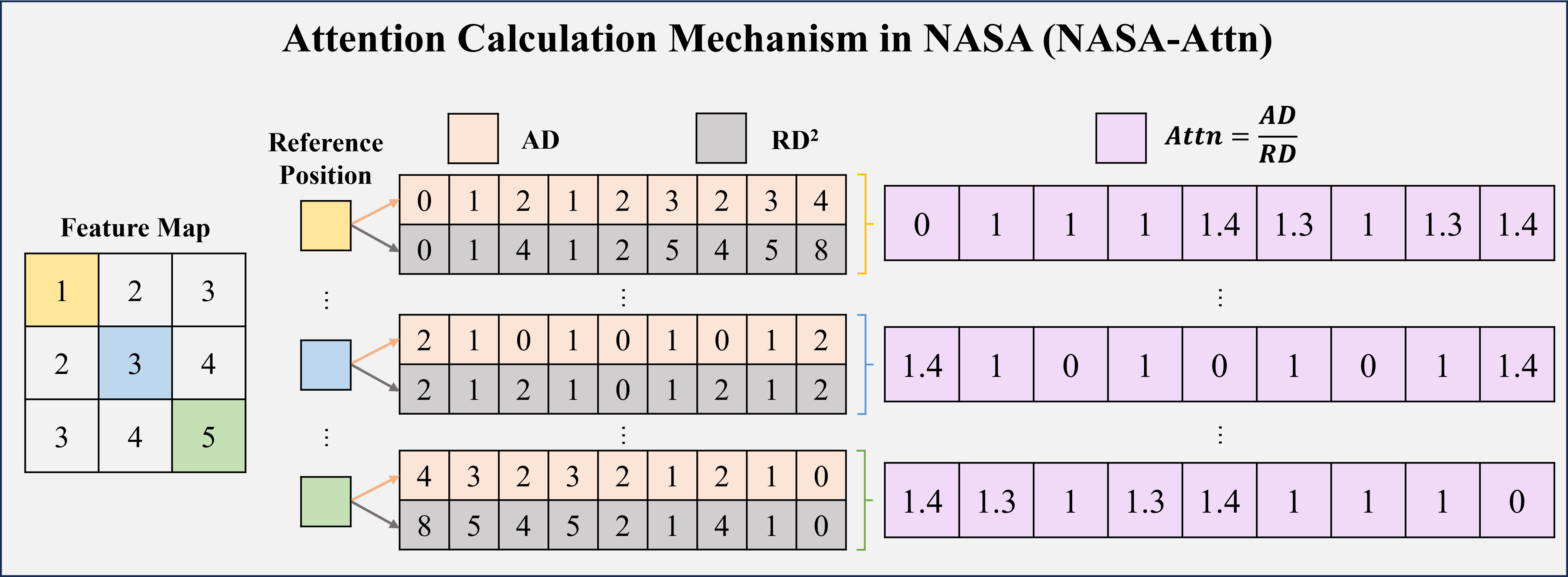}

   \caption{The attention calculation mechanism in NASA (\textbf{NASA-Attn}). \textbf{AD} and \textbf{RD} respectively denote the \textbf{A}bsolute \textbf{D}ifference and \textbf{R}elative \textbf{D}istance with respect to the reference pixel.}
   \label{fig:NASA-Attn-Calculation}
\end{figure}

\subsection{Noise-Aware Self-Attention Module} \label{sec:NASA}
To delve into potential artifacts arising from image noise residuals, we introduce a Noise-Aware Self-Attention \textbf{(NASA)} module. In contrast to conventional self-attention modules, we refrain from directly learning query and key matrices for attention calculation. Instead, we follow specific rules to craft an attention mechanism focusing more on the possible noise-related features.

To provide a clear illustration of NASA's design principle, we employ a simple example for clarification.
The image noise may lead to anomalous activation patterns (i.e. anomalous feature values) in feature maps.
As depicted in Fig~\ref{fig:NASA-Matrix-Interpre}, the two $4\times4$ matrices represent feature maps, with their elements denoting feature values. In noise-free regions, feature values exhibit smooth variations as shown in Fig~\ref{fig:NASA-normal-matrix}, while the presence of noise leads to dramatic feature value fluctuations as in Fig~\ref{fig:NASA-noise-matrix}.

However, simply identifying feature differences is not the optimal solution of searching for potential noise-related features. From the perspective of the self-attention mechanism, we utilize the variations of all features relative to the feature at position $(0,0)$ as example. In Fig~\ref{fig:NASA-noise-matrix}, we observe that the feature values at positions $(1,1)$ and $(3,3)$ both exhibit same and significant fluctuations. However, the underlying causes behind these fluctuations are distinct. The former is primarily attributed to noise interference, while the latter is related to disparities in image content due to the long relative distance. This highlights the importance of considering both feature value variations and the relative distance between features when identifying potential noise-related features based on a reference point. From the viewpoint of the reference point, positions with larger feature value variations and shorter relative distances should receive greater attention.

Inspired by this insight, we devise a novel attention calculation method. In the conventional self-attention mechanism, the element in the attention matrix at $(i,j)$ indicates the importance of the $j$th element to the $i$th element. In NASA, we redefine the element value in the attention matrix at $(i,j)$ as the likelihood that the $j$th element represents noise in relation to the $i$th element. As mentioned above, this likelihood is associated with the intensity of the variation and the spatial distance between two features. Consequently, in NASA, we reformulate the attention calculation mechanism (NASA-Attn) as follows:
\begin{equation}
    Attn_{(i,j)}=\frac{AD_{(i,j)}}{RD_{(i,j)}},
    % \frac{|f_j-f_i|}{dis(i,j)},
    \label{eq:NASA-calculation}
\end{equation}
where $AD_{(i,j)}$ and $RD_{(i,j)}$ represent the absolute difference and spatial relative distance between two features at the positions $i$ and $j$, respectively. Specifically, we define
\begin{equation}
    \begin{aligned}
        AD_{(i,j)}&=|f_i-f_j|, \\
        RD_{(i,j)}&= ||p_i-p_j||_2
        % RD_{(i,j)}&=\sqrt{(x_j-x_i)^2+(y_j-y_i)^2},
    \end{aligned}
    \label{eq:NASA-calculation-AD-RD}
\end{equation}
where $f_i$ and $f_j$ respectively represent the feature values at the position $i$ and $j$. $p_i=(x_i,y_i)$ and $p_j=(x_j,y_j)$ are the position coordinates of the $i$th and $j$th elements in the feature map.
Fig~\ref{fig:NASA-Attn-Calculation} provides a detailed view of NASA-Attn.

\begin{figure}[t]
  \centering
  % \fbox{\rule{0pt}{2in} \rule{0.9\linewidth}{0pt}}
   \includegraphics[width=\linewidth]{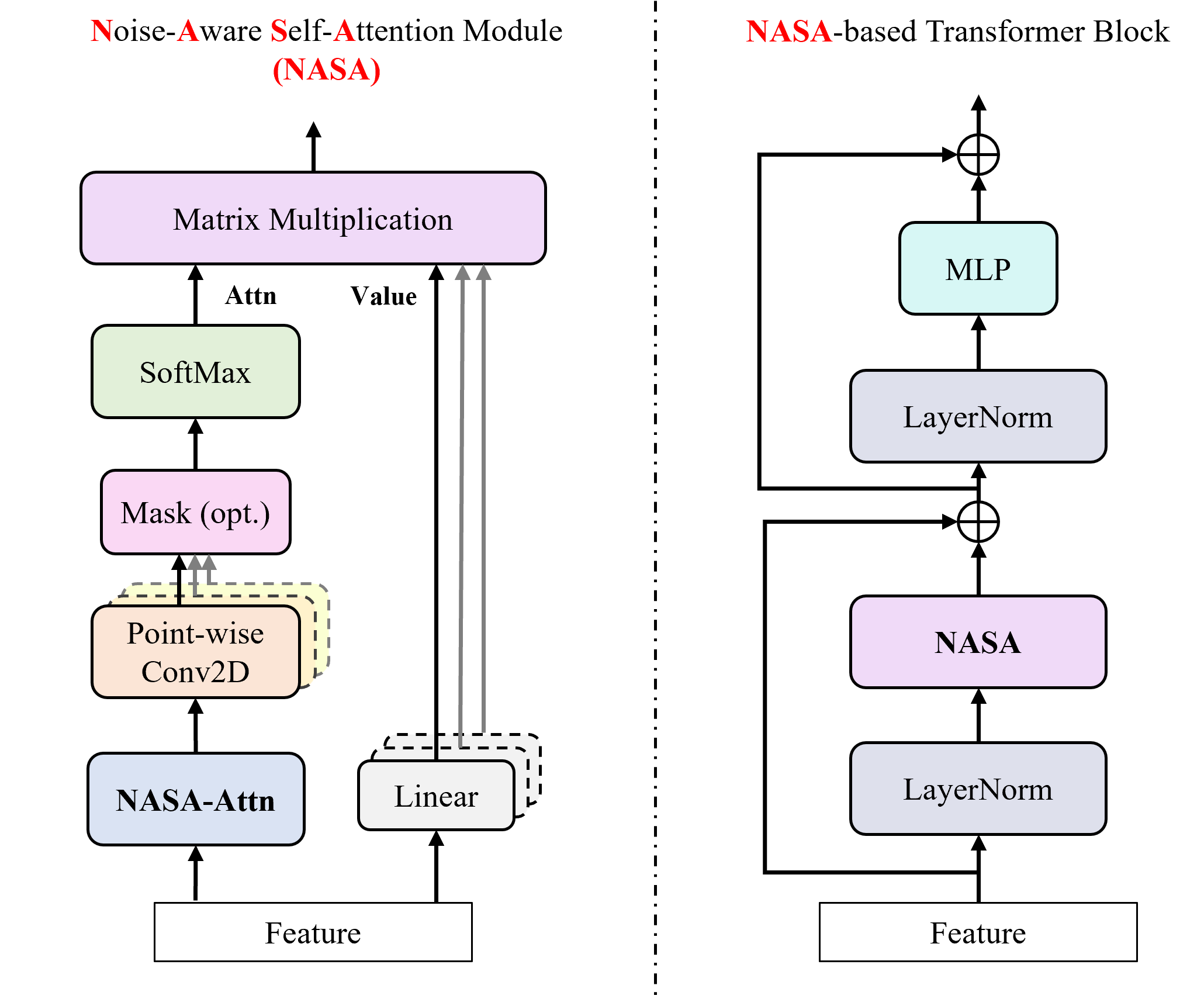}

   \caption{The architectures of NASA (left) and NASA-based Transformer Block (right). The dashed boxes show the multi-head NASA designs. The Mask module here is utilized to achieve the efficient batch computation for shifted configuration, which is proposed in Swin Transformer \cite{liu2021swin}.}
   \label{fig:NASA-Transformer-Block}
\end{figure}

Subsequently, to enhance the flexibility and adaptability of the self-attention mechanism, we introduce a point-wise convolution layer after the initial attention matrix calculated by NASA-Attn. The adjusted attention matrix is then passed through a softmax function to assign attention weights to each value. The remaining modules in the NASA-based Transformer Block resemble the conventional transformer architecture. For better understanding, please refer to Fig~\ref{fig:NASA-Transformer-Block}, which provides detailed information on NASA-based Transformer Block, including the multi-head version of NASA.

\begin{table*}
  \centering
  \caption{\textbf{Cross-domain evaluations.} We report Acc (\%) on the other forgery data generated by unseen generative models: ADM, Glide, Midjourney, VQDM, Wukong and BigGAN. The best results are in bold. And the second-best values are underlined. $\dag$ indicates that the results are provided by running their official codes. The results of LaRE$^2$ are directly cited from its official paper \cite{luo2024lare} and the other results are directly cited from \cite{zhu2023genimage}.}
  \resizebox{0.8\linewidth}{!}{
  \begin{tabular}{l c c c c c c c}
    \toprule
    \multirow{2}{*}{Method} & \multicolumn{6}{c}{Testing Subset} & \multirow{2}{*}{Avg-Acc}\\
    \cmidrule{2-7}
    & ADM & Glide & Midjourney & VQDM & Wukong & BigGAN \\
    \midrule
    ResNet-50 \cite{he2016deep} & $53.5$ & $61.9$ & $54.9$ & $56.6$ & $98.2$ & $52.0$ & $62.9$ \\
    DeiT-S \cite{touvron2021training} & $49.8$ & $58.1$ & $55.6$ & $56.9$ & $98.9$ & $53.5 $ & $62.1$ \\
    Swin-T \cite{liu2021swin} & $49.8$ & ${67.6}$ & ${62.1}$ & ${62.3}$ & ${99.1}$ & ${57.6}$ & ${66.4}$ \\
    {Swin-B \cite{liu2021swin} $\dag$} & ${55.6}$ & ${69.8}$ & ${54.1}$ & ${59.9}$ & ${99.8}$ & ${67.8}$ & ${67.8}$ \\
    \midrule
    F3Net \cite{qian2020thinking} & $49.9$ & $50.0$ & $50.1$ & $49.9$ & $\underline{99.9}$ & $49.9$ & $58.3$ \\
    GramNet \cite{liu2020global}  & $50.3$ & $54.6$ & $54.2$ & $50.8$ & $98.9$ & $51.7$ & $60.1$ \\
    CNNSpot \cite{wang2020cnn} & $50.1$ & $39.8$ & $52.8$ & $53.4$ & $78.6$ & $46.8$ & $53.6$ \\
    Spec \cite{zhang2019detecting} & $49.7$ & $ 49.8$ & $52.0$ & $55.6$ & $94.8$ & $49.8$ & $58.6$ \\
    Grag2021 \cite{gragnaniello2021gan} $\dag$ & $58.8$ & $75.5$ & $55.2$ & $72.3$ & $99.5$ & $\underline{74.8}$ & $72.7$ \\
    PatchForensics \cite{chai2020makes} $\dag$ & $63.0$ & $79.6$ & $65.5$ & $66.3$ & $94.8$ & $73.3$ & $73.8$ \\
    DIRE \cite{wang2023dire} $\dag$ & $\bm{75.1}$ & $76.3$ & $\bm{87.6}$ & $75.7$ & $95.7$ & $68.8$ & ${79.9}$ \\
    LaRE$^2$ \cite{luo2024lare} & $61.7$ & $\underline{88.5}$ & $74.0$ & $\bm{97.2}$ & $\bm{100.0}$ & $68.7$ & $\underline{81.7}$ \\
    \midrule
    \textbf{NASA-Swin (ours)} & $\underline{74.3}$ & $\bm{98.0}$ & $\underline{85.3}$ & $\underline{86.3}$ & $98.5$ & $\bm{93.0}$ & $\bm{89.2}$ \\
    \bottomrule
  \end{tabular}
  }
  \label{tab:cross-domain-eval}
\end{table*}

\subsection{NASA-Swin}
% Network
To effectively leverage NASA, we introduce NASA-Swin, an architecture designed for detecting diffusion-generated images, which incorporates Swin Transformer \cite{liu2021swin} as the backbone as shown in Fig~\ref{fig:architecture}. The window attention mechanism of Swin Transformer restricts attention matrix calculations to a local window. In the case of NASA-Swin, this confinement reduces feature value variations resulting from disparities in image content. Consequently, the window attention mechanism assists NASA in identifying potential noise-related features and distributing attention matrix weights sensibly across each point.

% Loss Function
Finally, as a binary classification task, we simply employ cross-entropy loss for model training. 

\section{Experiments}

\begin{table}
  \centering
  \caption{\textbf{In-domain evaluations.} We report Acc (\%) on Stable Diffusion V1.4 and V1.5 data of GenImage dataset. The best results are in bold. $\dag$ indicates that the results are provided by running their official codes. The results of LaRE$^2$ are directly cited from its official paper \cite{luo2024lare} and the other results are directly cited from \cite{zhu2023genimage}.}
  \resizebox{0.8\linewidth}{!}{
  \begin{tabular}{l c c}
    \toprule
    \multirow{2}{*}{Method} & \multicolumn{2}{c}{Testing Subset} \\
    \cmidrule{2-3}
    & SD V1.4 & SD V1.5 \\
    \midrule
    ResNet-50 \cite{he2016deep} & ${99.9}$ & $99.7$ \\
    DeiT-S \cite{touvron2021training} & ${99.9}$ & $99.8$ \\
    Swin-T \cite{liu2021swin} & ${99.9}$ & $99.8$ \\
    {Swin-B \cite{liu2021swin} $\dag$} & ${99.9}$ & ${\bm{99.9}}$ \\
    \midrule
    F3Net \cite{qian2020thinking} & ${99.9}$ & $\bm{99.9}$ \\
    GramNet \cite{liu2020global} & $99.2$ & $99.1$ \\
    CNNSpot \cite{wang2020cnn} & $96.3$ & $95.9$ \\
    Spec \cite{zhang2019detecting} & $99.4$ & $99.2$ \\
    Grag2021 \cite{gragnaniello2021gan} $\dag$ & ${99.9}$ & $\bm{99.9}$ \\
    PatchForensics \cite{chai2020makes} $\dag$ & $99.3$ & $99.3$ \\
    DIRE \cite{wang2023dire} $\dag$ & $96.2$ & $96.5$ \\
    LaRE$^2$ \cite{luo2024lare}  & $\bm{100.0}$ & $\bm{99.9}$ \\
    \midrule
    \textbf{NASA-Swin (ours)} & $99.0$ & $99.0$ \\
    \bottomrule
  \end{tabular}
  }
  \label{tab:in-domain-eval}
\end{table}

\subsection{Experimental Settings}
\textbf{Datasets.} In our experiments, we adopt the recently published large dataset, GenImage \cite{zhu2023genimage} as our data source. This dataset contains images generated by eight advanced generative models, consisting of seven diffusion-based models (i.e. ADM \cite{dhariwal2021diffusion}, Glide \cite{nichol2021glide}, VQDM \cite{gu2022vector}, Stable Diffusion V1.4 \cite{rombach2022high} (SD V1.4), Stable Diffusion V1.5 \cite{rombach2022high} (SD V1.5), Midjourney \cite{Midjourney} and Wukong \cite{Wukong}), along with a GAN-based model, BigGAN \cite{brock2018large}. GenImage effectively leverages the rich image content from ImageNet \cite{deng2009imagenet} to generate images containing various classes. In total, GenImage consists of $1,331,167$ real images (sampled from ImageNet) and $1,350,000$ synthetic images. Following the official settings, our training data exclusively consists of Stable Diffusion V1.4. It is noticed that SD V1.5, although fine-tuned with additional data, possesses an identical architecture to SD V1.4. Consequently, we utilize SD v1.4 and SD v1.5 as the in-domain evaluation datasets. To assess the generalization performance on unseen generative methods, we evaluate the trained models on synthetic images generated by the other generators. 

\begin{table*}
  \centering
  \caption{\textbf{Ablation study on various input modalities.} All models are trained \textbf{without CMS}. When dealing with the combination of RGB and noise inputs, we employ \textbf{CMFE} to fuse both modalities. We report Acc (\%) as the evaluation metric. The best results are in bold. And the second-best values are underlined.}
  \resizebox{0.8\linewidth}{!}{
  \begin{tabular}{c c c c c c c c c c}
    \toprule
    \multirow{2}{*}{Model} & \multicolumn{2}{c}{Modality} & \multicolumn{6}{c}{Testing Subset} & \multirow{2}{*}{Avg-Acc}\\
    \cmidrule(r){2-3} \cmidrule(r){4-9}
    & RGB & Noise & ADM & Glide & Midjourney & VQDM & Wukong & BigGAN & \\
    \midrule
    \multirow{3}{*}{{Swin-B \cite{liu2021swin}}} & $\checkmark$ & $-$ & ${55.6}$ & ${69.8}$ & ${54.1}$ & ${59.9}$ & ${\bm{99.8}}$ & ${67.8}$ & ${67.8}$ \\
    & $-$ & $\checkmark$ & $\bm{69.6}$ & $\underline{83.5}$ & $\bm{76.8}$ & $\bm{88.4}$ & $\underline{99.5}$ & $\underline{69.2}$ & $\underline{81.2}$ \\
    \cmidrule{2-10}
    & $\checkmark$ & $\checkmark$ & $\underline{60.2}$ & $\bm{94.0}$ & $\underline{74.5}$ & $\underline{81.2}$ & ${99.2}$ & $\bm{80.5}$ & $\bm{81.6}$ \\
    \midrule
    \midrule
    \multirow{3}{*}{\textbf{NASA-Swin}} & $\checkmark$ & $-$ & ${61.2}$ & ${83.5}$ & $\underline{74.0}$ & ${70.5}$ & $\underline{99.3}$ & ${67.9}$ & ${76.1}$ \\
    & $-$ & $\checkmark$ & $\bm{64.3}$ & $\underline{85.0}$ & $65.7$ & $\bm{81.3}$ & $\bm{99.7}$ & $\bm{87.8}$ & $\underline{80.6}$ \\
    \cmidrule{2-10}
    & $\checkmark$ & $\checkmark$ & $\underline{61.6}$ & $\bm{92.7}$ & $\bm{82.6}$ & $\underline{78.3}$ & $98.3$ & $\underline{79.0}$ & $\bm{82.1}$\\
    \bottomrule
  \end{tabular}
  }
  \label{tab:ablation-modality}
\end{table*}

\textbf{Baselines.} We evaluate the performance of our method by comparing it to several other detectors, including image classification networks such as ResNet-50 \cite{he2016deep}, DeiT-S \cite{touvron2021training}, and {Swin-T/B} \cite{liu2021swin}, forgery face detectors based on spatial (GramNet \cite{liu2020global}) and frequency (F3Net \cite{qian2020thinking}) artifacts, and general forgery image detectors trained with carefully-processed data (CNNSpot \cite{wang2020cnn}) and based on frequency clues (Spec \cite{zhang2019detecting}). 
Moreover, according to \cite{10095167}, we include PatchForensics \cite{chai2020makes} and Grag2021 \cite{gragnaniello2021gan} for comparison. While initially designed for GAN-generated image detection, both PatchForensics and Grag2021 exhibit good performance in the task of diffusion-generated image detection \cite{10095167}. Furthermore, we present the performance of DIRE \cite{wang2023dire} and LaRE$^2$ \cite{luo2024lare}, specifically designed for detecting diffusion-generated images.

\textbf{Evaluation Metrics.} In our experiments, we follow the evaluation settings of GenImage to report accuracy (Acc) as the evaluation metric. The threshold for computing accuracy is set to $0.5$, in line with \cite{wang2020cnn}. The experimental results of other detectors used for comparison are directly cited. 

\textbf{Implementation Details.} In our experiments, we initialize NASA-Swin using ImageNet pretrained weights. Notably, the NASA-Swin blocks share the same initial weights as the corresponding Swin blocks. Our model is trained for a maximum of $10$ epochs. The input images are center-cropped and resized to a size of $224\times224$. During training, we employ the SGD optimizer \cite{robbins1951stochastic} with a learning rate of $0.001$. The batch size is set to $64$, and no additional data augmentation techniques are utilized.

\subsection{Comparison to Existing Detectors}
In this section, we conduct a comparative analysis of our method against previous detectors, considering both in-domain and cross-domain scenarios. As established in prior studies \cite{wang2023dire, zhu2023genimage}, detectors trained on diffusion-generated images consistently demonstrate impressive performance within the confines of in-domain assessments. However, they often face challenges in terms of generalization when confronted with cross-domain evaluations. Our primary objective is to bolster the generalization capabilities of detectors when encountering images generated by previously unseen generative methods. To provide a comprehensive assessment, we present separate performance evaluations for both intra-domain and inter-domain datasets.

\textbf{Cross-domain Evaluations.} To begin, we assess the generalization capability of our method. Following the forgery detection protocol outlined in \cite{zhu2023genimage}, we train our NASA-Swin on SD V1.4 data and test it on forgery data generated by previously unseen diffusion models, including ADM, Glide, Midjourney, VQDM, and Wukong. Notably, we also evaluate our model on BigGAN-generated images, a representative method within the GAN family \cite{zhu2023genimage}. This presents a challenging cross-domain evaluation due to the distinct generative process compared to diffusion models.

Table~\ref{tab:cross-domain-eval} displays the results of our cross-domain evaluations. Previous forgery detectors tend to struggle when detecting images generated by previously unseen generative methods. Surprisingly, some detectors (CNNSpot \cite{wang2020cnn} and Spec \cite{zhang2019detecting}) specifically designed for GAN-generated images perform even worse compared to binary image classifiers. While the advanced binary classifier, Swin Transformer, with its strong representation capability, exhibits better performance. This disparity may be attributed to the differences in generative processes between GAN and diffusion models, rendering previous detection clues less suitable for identifying diffusion-generated images. 
Additionally, although DIRE \cite{wang2023dire} and LaRE$^2$ \cite{luo2024lare} exhibit significant performance improvements on other unseen diffusion-generated images, they still show limited detection performance on BigGAN-generated images.
These observations have prompted us to develop novel forgery clues for effective detection.

NASA-Swin achieves state-of-the-art results when detecting forgery images generated by previously unseen diffusion models. Notably, it excels even when faced with BigGAN-generated images 
({Swin-B vs. NASA-Swin: $67.8\%$ vs. $93.0\%$}).
NASA-Swin outperforms its baseline ({Swin-B}) by a considerable margin on these images, with an average accuracy (Avg-Acc) improvement of {$21.4\%$}. 
Furthermore, compared to DIRE and LaRE$^2$, NASA-Swin shows superior detection performance (Avg-Acc: NASA-Swin vs. DIRE/LaRE$^2$ $\rightarrow$ $89.2\%$ vs. $79.9\%/81.7\%$). Particularly, NASA-Swin exhibits significant improvements on BigGAN-generated images (from $68.8\%/68.7\%$ to $93.0\%$).
These results highlight the outstanding performance of our NASA-Swin when encountering previously unseen generative methods, demonstrating its superior generalization capability compared to previous detectors, even when confronted with generators employing different generative processes (BigGAN).

\begin{table*}
  \centering
  \caption{\textbf{Ablation study on different architectures of Noise-Aware Block.} We report Acc (\%) as the evaluation metric. The best results are in bold. And the second-best values are underlined.}
  \resizebox{0.7\linewidth}{!}{
  \begin{tabular}{c c c c c c c c}
    \toprule
    \multirow{2}{*}{\makecell[c]{Noise-Aware\\Block}} & \multicolumn{6}{c}{Testing Subset} & \multirow{2}{*}{Avg-Acc}\\
    \cmidrule{2-7}
    & ADM & Glide & Midjourney & VQDM & Wukong & BigGAN & \\
    \midrule
    \XSolidBrush & $65.2$ & $97.4$ & $76.2$ & $76.2$ & $\bm{99.0}$ & $82.4$ & $82.7$ \\
    Swin & $\underline{69.3}$ & $\bm{98.1}$ & $\underline{81.9}$ & $\underline{76.3}$ & $\underline{98.7}$ & $\underline{84.5}$ & $\underline{84.8}$ \\
    NASA & $\bm{74.3}$ & $\underline{98.0}$ & $\bm{85.3}$ & $\bm{86.3}$ & $98.5$ & $\bm{93.0}$ & $\bm{89.2}$ \\
    \bottomrule
  \end{tabular}
  }
  \label{tab:ablation-NASA}
\end{table*}

\begin{table*}
    \centering
    \caption{\textbf{Ablation study of relative distance term in attention calculation mechanism within NASA.} We report Acc (\%) as the evaluation metric. The best results are in bold.}
    \begin{tabular}{c c c c c c c c c}
    \toprule
    \multirow{2}{*}{\makecell[c]{Method}} & \multirow{2}{*}{\makecell[c]{RD}} & \multicolumn{6}{c}{Testing Subset} & \multirow{2}{*}{Avg-Acc}\\
    \cmidrule{3-8}
    &  & ADM & Glide & Midjourney & VQDM & Wukong & BigGAN & \\
    \midrule
    NASA-Swin & w/o & $65.9$ & $94.9$ & $80.1$ & $84.8$ & $\bf{99.2}$ & $74.1$ & $83.2$ \\
    NASA-Swin & w & $\bf{74.3}$ & $\bf{98.0}$ & $\bf{85.3}$ & $\bf{86.3}$ & $98.5$ & $\bf{93.0}$ & $\bf{89.2}$ \\
    \bottomrule
    \end{tabular}
    \label{tab:ablation-RD}
\end{table*}

\begin{table*}
  \centering
  \caption{\textbf{Ablation study on all components of NASA-Swin.} {Swin-B*} directly employs the concatenation of noise residual and RGB image pairs as input without any other fusion strategy. We report Acc (\%) as the evaluation metric. The best results are in bold. And the second-best values are underlined.}
  \small
  \begin{tabular}{c c c c c c c c c c c c}
    \toprule
    \multirow{2}{*}{Model} & \multirow{2}{*}{Modality} & \multirow{2}{*}{CMFE} & \multirow{2}{*}{CMS} & \multirow{2}{*}{NASA} & \multicolumn{6}{c }{Testing Subset} & \multirow{2}{*}{Avg-Acc}\\
    \cmidrule{6-11}
    &  &  &  &  & ADM & Glide & Midjourney & VQDM & Wukong & BigGAN & \\
    \midrule
    {Swin-B \cite{liu2021swin}} & RGB & $-$ & $-$ & $-$ & ${55.6}$ & ${69.8}$ & ${54.1}$ & ${59.9}$ & ${\bm{99.8}}$ & ${67.8}$ & ${67.8}$ \\
    {Swin-B* \cite{liu2021swin}} & Both & $-$ & $-$ & $-$ & $52.4$ & $66.3$ & $59.6$ & $51.6$ & $99.0$ & $54.3$ & $63.9$ \\
    {Swin-B \cite{liu2021swin}} & Both & $\checkmark$ & $-$ & $-$ & $60.2$ & ${94.0}$ & $74.5$ & $\underline{81.2}$ & $\underline{99.2}$ & ${80.5}$ & ${81.6}$ \\
    {Swin-B \cite{liu2021swin}} & Both & $\checkmark$ & $\checkmark$ & $-$ & $\underline{65.2}$ & $\underline{97.4}$ & ${76.2}$ & $76.2$ & $99.0$ & $\underline{82.4}$ & $\underline{82.7}$ \\
    NASA-Swin & Both & $\checkmark$ & $-$ & $\checkmark$ & $61.6$ & $92.7$ & $\underline{82.6}$ & $78.3$ & $98.3$ & $79.0$ & $82.1$ \\
    NASA-Swin & Both & $\checkmark$ & $\checkmark$ & $\checkmark$ &  $\bm{74.3}$ & $\bm{98.0}$ & $\bm{85.3}$ & $\bm{86.3}$ & $98.5$ & $\bm{93.0}$ & $\bm{89.2}$ \\
    \bottomrule
  \end{tabular}
  \label{tab:ablation-all-components}
\end{table*}

\textbf{In-domain Evaluations.} In addition to the cross-domain comparisons, we also present the detection results on SD v1.4 and SD v1.5 datasets from GenImage, which serve as the in-domain evaluation. The results are detailed in Table~\ref{tab:in-domain-eval}. Previous approaches that leverage binary image classifiers typically achieve remarkable performance in in-domain assessments 
(e.g., {Swin-T/B \cite{liu2021swin} achieves $99.9\%/99.9\%$ and $99.8\%/99.9\%$ on SD V1.4 and SD V1.5, respectively}).
Conversely, many methods employing well-designed architectures and training strategies compel models to learn specific forgery artifacts within generated images, which slightly restricts model performance in in-domain evaluations (e.g., GramNet \cite{liu2020global} achieving $99.2\%$ and $99.1\%$ on SD V1.4 and SD V1.5, respectively). Notably, F3Net \cite{qian2020thinking} exhibits outstanding performance, suggesting the effectiveness of frequency information. In comparison to the aforementioned methods, our approach incorporates residual noise information into the detection framework. While our performance experiences a slight decrease, it remains competitive with others ($99.0\%$ and $99.0\%$ on SD V1.4 and SD V1.5, respectively).

\begin{figure*}[t]
  \centering
   \includegraphics[width=\linewidth]{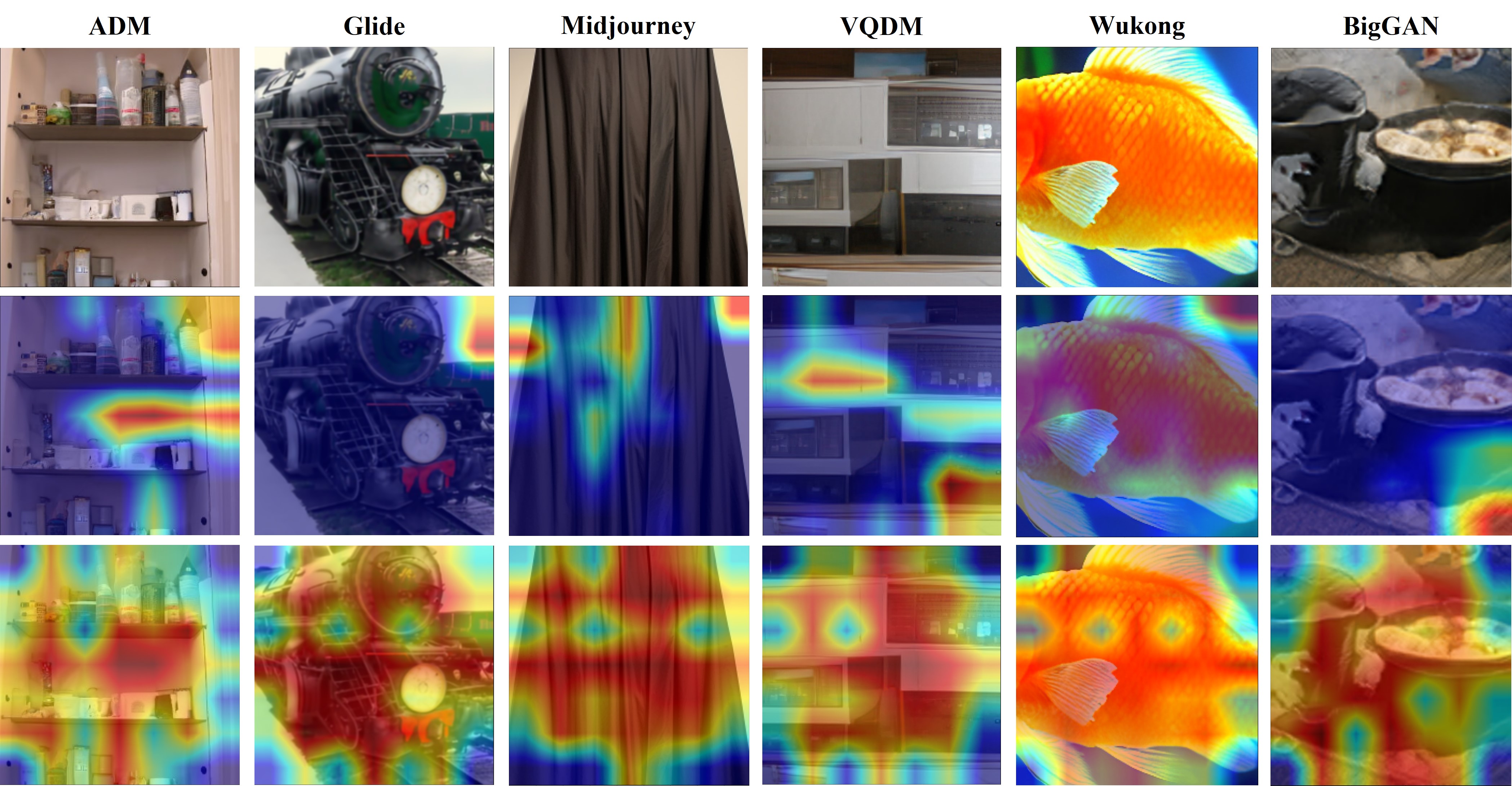}

   \caption{Saliency map visualization of Swin-Transformer and NASA-Swin detectors on the images generated by different generative models. The first row is the forgery images synthesized by different generative models. The middle and last rows are respectively the saliency maps of Swin-Transformer and NASA-Swin detectors on these images.}
   \label{fig:CAM-Comparison}
\end{figure*}

\begin{table*}
  \centering
  \caption{\textbf{Analysis of NASA-Swin variant architectures by incorporating NASA-based branch in different stages.} We report Acc (\%) as the evaluation metric. The best results are in bold. And the second-best values are underlined. Avg-Acc denotes the average accuracy among various inter-domain forgery data.}
  \label{tab:ablation-NASA-branch}
  \begin{tabular}{c c c c c c c c c c}
    \toprule
    \multirow{2}{*}{\makecell[c]{NASA-based\\Branch Stages}} & \multicolumn{2}{c}{In-Domain Testing} & \multicolumn{7}{c}{Cross-Domain Testing}\\
    \cmidrule(r){2-3} \cmidrule{4-10}
    & SD V1.4 & SD V1.5 & ADM & Glide & Midjourney & VQDM & Wukong & BigGAN & Avg-Acc \\
    \midrule
    1-2 & $\cellcolor{lightgray}99.7$ & $\cellcolor{lightgray}99.8$ & $60.0$ & $94.0$ & $62.9$ & $73.4$ & $\bm{99.7}$ & $82.8$ & $78.8$ \\
    \textbf{2-3} & $\cellcolor{lightgray}99.0$ & $\cellcolor{lightgray}99.0$ & $\underline{74.3}$ & $\bm{98.0}$ & $\bm{85.3}$ & $\underline{86.3}$ & $98.5$ & $\bm{93.0}$ & $\bm{89.2}$ \\
    3-4 & $\cellcolor{lightgray}99.1$ & $\cellcolor{lightgray}99.0$ & ${73.3}$ & $\bm{98.0}$ & ${84.1}$ & $\bm{88.8}$ & $98.8$ & $\underline{87.5}$ & $\underline{88.4}$ \\
    {1-3} & $\cellcolor{lightgray}{99.3}$ & $\cellcolor{lightgray}{99.2}$ & ${\bm{74.9}}$ & ${\underline{97.3}}$ & ${\underline{84.7}}$ & ${84.8}$ & ${99.0}$ & ${79.3}$ & ${86.7}$ \\
    {2-4} & $\cellcolor{lightgray}{99.6}$ & $\cellcolor{lightgray}{99.6}$ & ${62.6}$ & ${96.1}$ & ${73.3}$ & ${74.4}$ & ${\underline{99.3}}$ & ${84.6}$ & ${81.7}$ \\
    All & $\cellcolor{lightgray}99.5$ & $\cellcolor{lightgray}99.4$ & $69.9$ & ${95.8}$ & $79.6$ & $80.3$ & ${99.2}$ & $81.5$ & $84.4$ \\
    \bottomrule
  \end{tabular}
\end{table*}

\subsection{Ablation Study}
In this section, we perform the ablation study to analyze the effects of each component in NASA-Swin. All experiments are trained on SD V1.4, and tested on inter-domain datasets.

\textbf{Study on Input Modality.} Our method incorporates noise residuals into RGB images, as we believe that the noise residuals 
exhibit inconsistent patterns between genuine and generated images, which can enhance detection performance.
To validate the effectiveness of this approach, we separately train Swin Transformer models using RGB images, noise residual images, and a combination of both. When using both modalities, we employ the proposed CMFE to handle RGB and noise residual image pairs. The comparison results are presented in Table~\ref{tab:ablation-modality}. 
We observe that when using single-modality inputs, noise residual images outperform RGB images in Avg-Acc, with noise residuals achieving $81.2\%$ compared to {$67.8\%$} for RGB. This demonstrates the effectiveness of noise residuals in forgery detection. 
When utilizing CMFE to fuse both modalities as input, detection performances are further improved in terms of Avg-Acc (from {$67.8\%$} and $81.2\%$ to $81.6\%$). 
Especially, on Glide and BigGAN images, the detection performances increase from {$69.8\%/83.5\%$} and {$67.8\%/69.2\%$} to $94.0\%$ and $80.5\%$, respectively.
These indicate that the detection clues from the two modalities can be complementary. Combining the two modalities proves beneficial in achieving enhanced generalization performance. 

However, it is worth noting that image noise is inherently present in the original (RGB) images. Our designed NASA-Swin enhances the focus on the noise-related information by an additional NASA-based branch. Therefore, we also would like to explore whether including image noise as an input modality is beneficial for our NASA-Swin. 
We compare the detection performances by respectively utilizing RGB images, noise images and a combination of both as the inputs to NASA-Swin. 
As shown in Table~\ref{tab:ablation-modality}, when using RGB images as inputs, NASA-Swin achieves a better performance in Avg-Acc than Swin Transformer ($76.1\%$ vs. {$67.8\%$}).
This demonstrates that our NASA-Swin can effectively capture the noise-related information concealed in RGB images. 
Additionally, it is noticed that compared to RGB inputs, noise inputs perform better in terms of Avg-Acc (RGB inputs vs. Noise inputs: $76.1\%$ vs. $80.6\%$). 
{Especially, due to directly learning forgery patterns from noise residual images, Swin Transformer shows comparable and even slightly superior detection performance to NASA-Swin with noise inputs in terms of Avg-Acc ($81.2\%$ and $80.6\%$). These results demonstrate the effectiveness of noise images on the generated image detection task.}
By combining both modalities in the input phase, the performance exhibits further improvement in terms of Avg-Acc, increasing from $76.1\%/80.6\%$ to $82.1\%$. This indicates that the detection clues from the two modality inputs may also complement each other in NASA-Swin.
It is evident that compared to the single RGB image inputs, the inclusion of noise inputs indeed contributes to NASA-Swin’s enhanced forgery detection performance.

\textbf{Effectiveness of Noise-Aware Self-Attention (NASA).}
In our architecture, NASA is a well-designed self-attention module utilized for capturing noise-related information to enhance detection performance. To validate its effectiveness, we conduct a comprehensive assessment focusing on three crucial aspects: \textbf{the importance of an additional noise-aware branch}, \textbf{the importance of NASA blocks in the noise-aware branch} and \textbf{the importance of incorporating relative distance into the attention calculation mechanism within NASA blocks}.

First, we assess the importance of adding an extra branch dedicated to handling noise-related information. To this end, we remove the noise-aware branch from our architecture, reverting it to a standard Swin Transformer (the x in Table~\ref{tab:ablation-NASA}). As shown in Table~\ref{tab:ablation-NASA}, our method, which employs NASA to additionally capture noise-related features, exhibit a significant improvement in detection performance (Avg-Acc increases from $82.7\%$ to $89.2\%$). Specifically, these enhancement is particular pronounced on the images generated by various generative methods, such as ADM ($+9.1\%$), Midjourney ($+9.1\%$), VQDM ($+10.1\%$), and BigGAN ($+10.6\%$). 
{Although our method has some slight performance decrease on Wukong subset ($-0.5\%$), it is still comparable to the standard Swin Transformer.}
These results convincingly support the necessity of the additional branch to effectively capture noise-related information and improve detection performance.

\begin{figure}
    \centering
    \subfloat[\small Feature Space of Swin-Transformer]{\includegraphics[width=\linewidth]{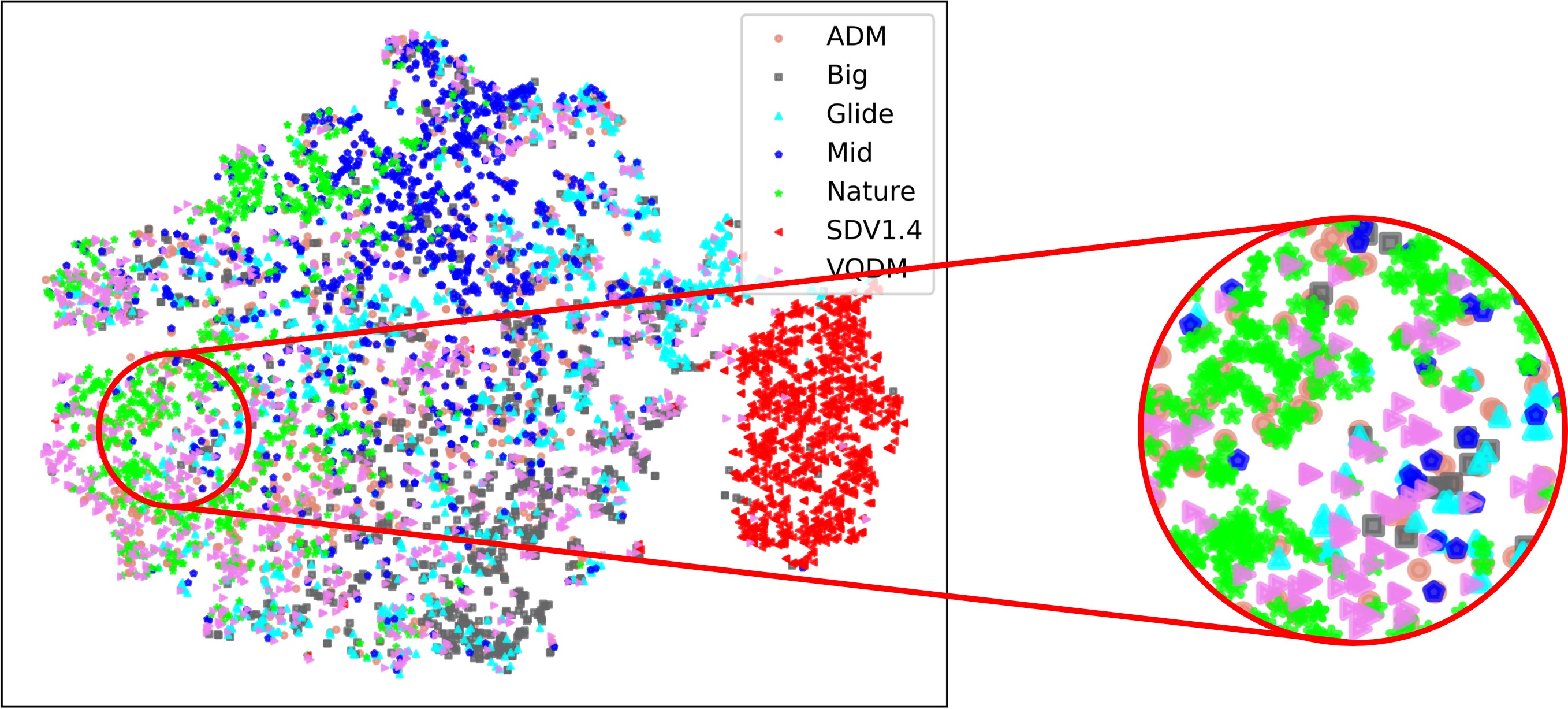}
    \label{fig:SwinBase-Feature-Space}}
    \hfill
    \subfloat[\small Feature Space of NASA-Swin]{\includegraphics[width=\linewidth]{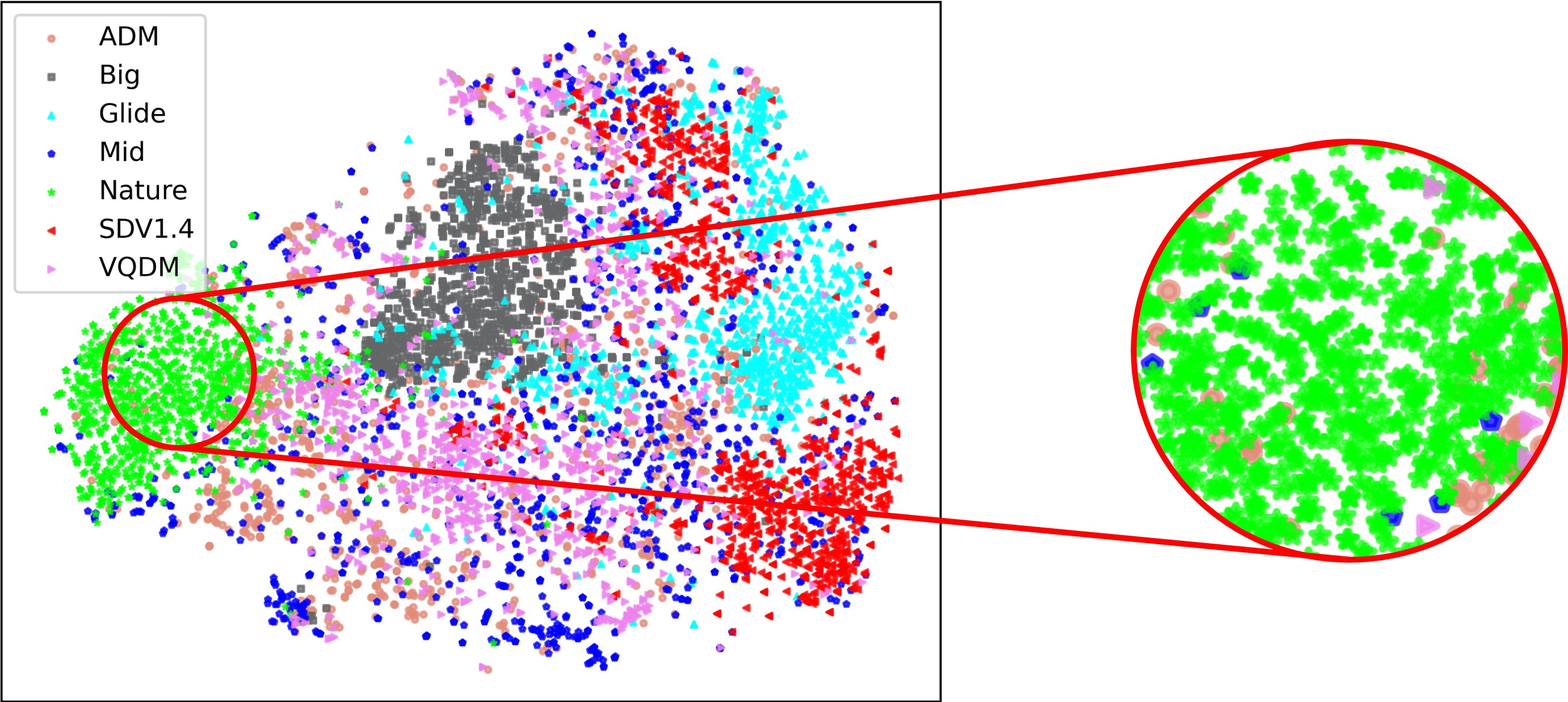}
    \label{fig:NASA-Feature-Space}}
    \hfill
    \caption{Feature space visualization of Swin-Transformer (a) and our NASA-Swin (b). The baseline detector, Swin-Transformer, cannot distinguish real images (i.e. Nature) from the forgery images generated by unseen methods (because the feature vectors fall into the same feature space), while our NASA-Swin succeeds in distinguishing real images from the synthesized images. Best viewed in color.}
    \label{fig:Feature-Space-Visualization}
\end{figure}

In the second aspect, we aim to investigate whether NASA blocks, which constitute the additional branch, are essential. To explore this, we replace the NASA blocks with standard Swin Transformer blocks for comparison. The results, as presented in Table~\ref{tab:ablation-NASA}, reveal that incorporating an additional branch composed of standard Swin Transformer blocks leads to only marginal improvement in detection performance (Avg-Acc increased from $82.7\%$ to $84.8\%$). This slight improvement might even be attributed to the increased complexity of the network architecture. Compared to it, our method, which leverages NASA blocks to capture noise-related information, exhibit impressive superiority on performance (Avg-Acc: $84.8\%$ vs. $89.2\%$). Specifically, the detection performance on unseen generative models shows notable advantages (e.g., ADM: $69.3\%$ vs. $74.3\%$ and VQDM: $76.3\%$ vs. $86.3\%$). 
{Although our method has some subtle performance fluctuations on the Glide subset ($-0.1\%$), it is still saturated (Acc: $98.0\%$)}.
These results demonstrate the necessity of the NASA block in the noise-aware branch, and its capacity to achieve superior detection performance.

Finally, we evaluate the importance of incorporating relative distance into the attention calculation mechanism with NASA blocks. To this end, we compare results with and without the relative distance term in the attention calculation procedure, as shown in Table~\ref{tab:ablation-RD}. By incorporating relative distance into NASA blocks, our method provides obvious improvements in generalization on unseen generative models (Avg-Acc: from $83.2\%$ to $89.2\%$). Particularly, the detection performance on BigGAN shows a significant improvement (from $74.1\%$ to $93.0\%$). This demonstrates that the relative distance plays a crucial role in improving generalization ability of forgery detectors, especially for the different generative paradigm.

\textbf{Effectiveness of CMFE and CMS.} In addtion to NASA, we also explore the influence of the proposed Cross-Modality Fusion Embedding (CMFE) module and Channel Mask Strategy (CMS) on detection performance. 

CMFE is specifically designed to fuse RGB and noise residual image pairs as input. To demonstrate its impact on detection performance, we compare it with a straightforward modality fusion strategy. In this strategy, we directly concatenate the noise residual and RGB image pairs as input and train a Swin Transformer. The results are presented in Table~\ref{tab:ablation-all-components}. The simple modality fusion strategy proves ineffective in improving performance and may even impair it. In contrast, our CMFE significantly outperforms the baseline on these data, with an average accuracy increase of {$13.8\%$}. This reveals the importance of the image modality fusion method, and validates the effectiveness of our CMFE.

CMS, as a data augmentation strategy, is applied to the input channels. As shown in Table~\ref{tab:ablation-all-components}, we conduct separate evaluations of Swin Transformer and NASA-Swin, with and without CMS. When equipped with CMS, the detection performance (Avg-Acc) of the trained models exhibits noticeable improvement (Swin Transformer: $1.1\%$, NASA-Swin: $7.1\%$). This clearly illustrates the effectiveness of CMS in enhancing the detection of generated images.

\subsection{Analysis}
\textbf{More NASA-Swin design variations.} In our implementations, we construct our NASA-Swin by extending an additional NASA-based branch in the intermediate stages (specifically, stage 2 and 3 in our implementation). The first stage is employed to learn unified and informative representations, while the last stage is utilized to integrate information from both branches and yield the final decision. Here, we conduct experiments to explore more possible architecture designs for NASA-Swin by incorporating the NASA-based branch in various stages. In addition to the standard configurations, we develop {five} additional designs for comparison 
{(extending NASA-based branch in stage 1 and 2, stage 3 and 4, stage 1-3, stage 2-4, and throughout all stages).}
The results are shown in Table~\ref{tab:ablation-NASA-branch}. From these comparisons, we observe that incorporating NASA-based branch in stage 2 and 3 has the best generalization ability on unseen generative models. This suggests that simultaneously ensuring the information integration process of the two modality inputs and the two feature branches may help the model improve the generalization performance.

\textbf{Analysis of saliency map visualization.} We utilize Grad-CAM \cite{selvaraju2017grad} to visualize the regions that forgery detectors focus on when analyzing synthesized images generated by unseen forgery models (including ADM, Glide, Midjourney, VQDM, Wukong and BigGAN). As shown in Fig~\ref{fig:CAM-Comparison}, compared to the baseline (i.e. Swin-Transformer) detector, NASA-Swin focuses more on the regions with rich contents and textures. This is because these regions generally contain rich noise-related information, which NASA-Swin can effectively capture to detect forgery images. As mentioned above (refer to Fig~\ref{fig:RID-Residual-FFT}), the noise-related information contributes to explore consistent forgery artifacts among various generative models. This may explain why our model improves the generalization performance of forgery detection.

\textbf{Analysis of feature space visualization.} We then utilize t-sne \cite{van2008visualizing} to visualize feature vectors from the last layers of the detection models, as shown in Fig~\ref{fig:Feature-Space-Visualization}. We would like to emphasize that both the baseline (Swin-Transformer) and our NASA-Swin models can effectively distinguish genuine and SD V1.4 images, because they are seen in the training phase. However, for the unseen forgery images, the baseline model struggles to distinguish them from real images, while our NASA-Swin exhibits significant improvements in recognizing the genuine and synthesized images. In addition, compared to the baseline, NASA-Swin demonstrates a broader overlap in the feature spaces of SD V1.4 and unseen forgery images. These results indicate that NASA-Swin enhance its capability to learn general forgery artifacts to improve the generalization performance of forgery detection.

\section{Conclusion}
This paper introduces a novel perspective for diffusion-generated image detection, leveraging the image residual noise to enhance generalization. 
We explicitly incorporate these image noise residuals into the input and employ a cross-modality fusion module to combine RGB and noise images. Additionally, we introduce a Noise-Aware Self-Attention mechanism, offering a unique attention calculation method. A new baseline for diffusion-generated image detection is provided, called NASA-Swin, which effectively captures noise-related information. We also introduce a channel mask strategy to encourage the model to extract more informative features. Comprehensive experiments demonstrate the effectiveness of each module, and comparisons with competing methods highlight NASA-Swin's superior generalization ability.

\bibliographystyle{ieeetr}
\bibliography{egbib}

\vfill

\vspace{-10 mm}

\end{document}